\documentclass{jov_arxiv}

\usepackage{amsmath,amssymb}
\usepackage{mathrsfs}
\usepackage{mathbbol}

\usepackage{array}
\usepackage{rotating}

\newcommand{\blue}[1]{\textcolor[rgb]{0,0,1}{#1}}

\newlength\savedwidth

\usepackage[all]{xy}
\usepackage{amsmath}
\usepackage{dsfont}
\usepackage{cancel}
\usepackage{lineno}
\usepackage{pbox}
\usepackage{bbm}
\begin{document}
\setpagewiselinenumbers
\modulolinenumbers[1]

\title{Paraphrasing Magritte's Observation}

\abstract{Contrast Sensitivity of the human visual system can be explained from certain low-level vision tasks (like retinal noise and optical blur removal), but not from others (like chromatic adaptation or pure reconstruction after simple bottlenecks). This conclusion still holds even under substantial change in stimulus statistics, as for instance considering cartoon-like images as opposed to natural images~\cite{Li22}.

In this note we present a method to generate original cartoon-like images compatible with the statistical training used in~\cite{Li22}. Following the classical observation in~\cite{Magritte29}, the stimuli generated by the proposed method certainly \emph{are not} what they represent: \emph{Ceci n'est pas une pipe}. The clear distinction between representation (the stimuli generated by the proposed method) and reality (the actual object) avoids eventual problems for the use of the generated stimuli in academic, non-profit, publications.
}

\author{Malo}{Jes\'us}
 {Image Processing Lab, Parc Cientific}
 {Universitat de València, Spain}
 {http://isp.uv.es}
 {jesus.malo@uv.es}


\maketitle

\section{1. Introduction}

Visual neuroscience based on statistical learning depends on image datasets that simulate the environment that shapes the sensors.
Seminal contributions were based on clever observations on few images~\cite{Field87}, and 2nd-order trends and marginal PDFs on decorrelated domains can be fairly obtained just from hundreds of images~\cite{Malo00}.
Specific effects such as chromatic adaptation require images with accurate calibration~\cite{Gutmann14,Laparra15,Gomez19,Malo20}. In this context small datasets with radiometrically controlled spectra~\cite{Foster15,Foster16}, or tristimulus calibrated images~\cite{Laparra12} are fundamental.
However, conventional deep-learning is data-intensive and hence larger datasets (with millions of images) are required~\cite{Russakovsky2015}.

Some academic journals require authors to inform about the copyright status of the images shown in the papers. In this setting, large datasets of images automatically taken from the internet are great for training models but they are problematic in terms of copyright or data protection: the database maybe publicly available for academic purposes but the specific author of each picture may be unknown. Moreover, if the picture displays people, they should sign the explicit consent to appear in the publication.
As a result, in practice it is not easy to display these images in scientific publications.

In case of natural images, endless arguments with editorial staff are easy to avoid by using whatever picture of the author taken by a close friend, as in~\cite{Martinez18}, or whatever picture of a close friend taken by the author, as in~\cite{Webster11}. In these situations obtaining the permissions is straightforward.

The solution may not be that obvious in the case of cartoon-like images. In this note we present a method to generate original cartoon-like images compatible with the statistical training used in~\cite{Li22}.

\section{2. Method}

In order to generate original cartoon-like images ready to be published we propose the following method: (1) deep introspection to internally visualize the main features of the class of scenes one wants to represent, and (2) by-hand drawing to synthesize an image of such mental representation.

\begin{figure}[b!]
\begin{center}
\includegraphics[width=0.95\textwidth]{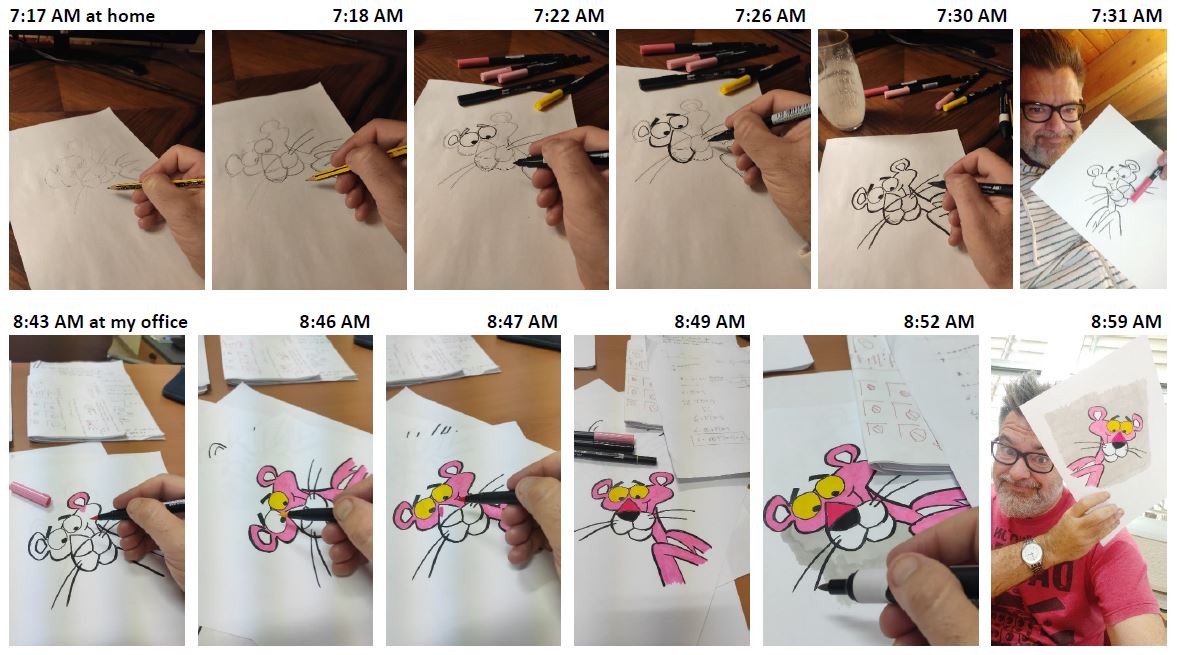}\\[-0.5cm]
\caption{By-hand drawing stage of the proposed method. Performance by the author executed by feb. 8th 2022 in the case of a popular cartoon. He took pictures of himself along the process to illustrate intermediate results.
The author holds the copyright of these pictures and explicitly consents that the pictures that display his face appear in this publication.}
\label{data}
\end{center}
\end{figure}

Figure \ref{data} illustrates the timeline and intermediate results of the by-hand drawing stage of the proposed method. While waking up that early or coffee is not strictly necessary, a bit of background in drawing may help to keep the image statistics right so that the synthesized signal \emph{resembles} the target.

It is important to note that given the randomness associated to the internal representation of stage (1) and the noisy nature of stage (2), the proposed synthesis method is unlikely to generate nothing close (in Mean Squared Error terms) to any existing still image of the target class.
More generally, following the classical observation in~\cite{Magritte29}, \emph{ceci n'est pas une pipe}, the stimuli generated by the proposed method certainly \emph{are not} the objects they represent.

\section{3. Results}

\begin{figure}[t!]
\begin{center}
\includegraphics[width=5.5cm]{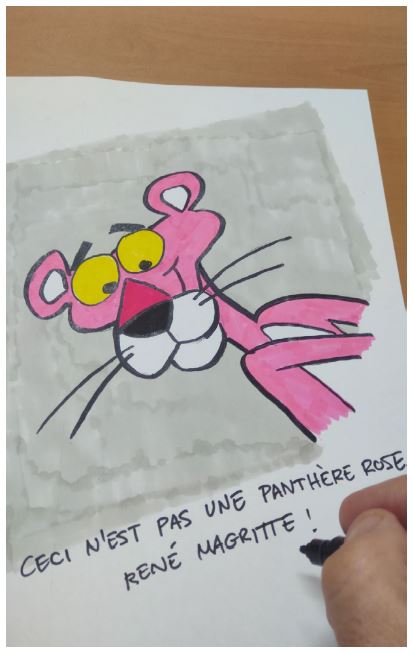}\hspace{1.5cm}\includegraphics[width=11cm]{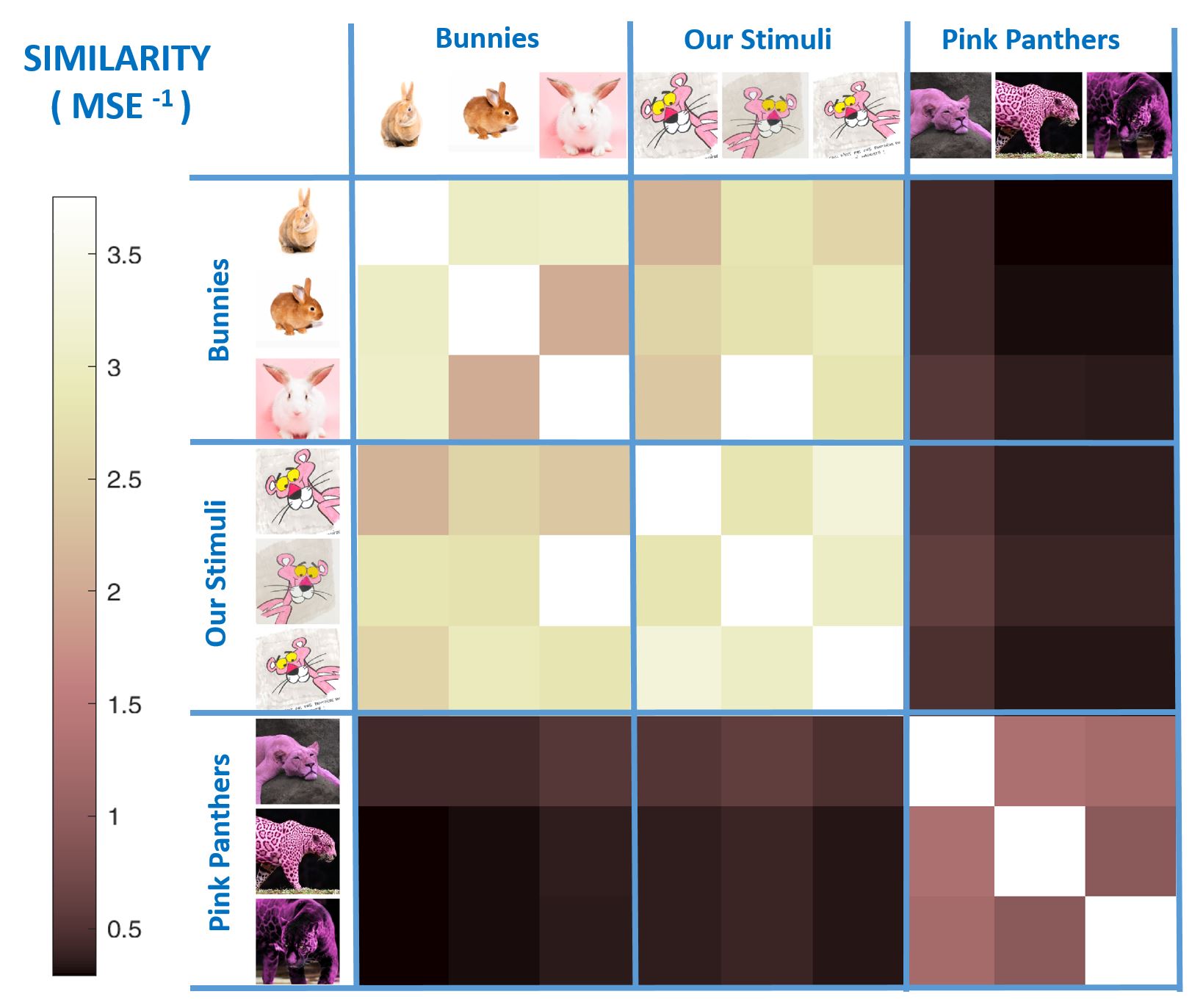}\\[-0.1cm]
\caption{\textbf{(Left)} Result of the proposed method together with a reference to~\blue{(Magritte, 1929)}. This image depicting the result of the proposed method was also taken by the author and he supports its publication here as well as any use of this image for academic non-profit purposes.
In order to compare this result with the original result of René Magritte, please visit https://en.wikipedia.org/wiki/File:MagrittePipe.jpg. \textbf{(Right)} Similarities based on the Mean Squared Error (MSE$^{-1}$) between the generated stimuli, bunnies and pink panthers. Lighter colors stand for bigger similarity.}
\label{main_result}
\end{center}
\end{figure}

Fig.~\ref{main_result} (left) shows the result of the proposed method. Note that the reflection of the images introduced by the careless use of the \emph{selfie} mode of the cellphone alters the appearance of time in the watch of the picture taken at 8:59AM in Fig.~\ref{data}, but it does not reduce the generality of the result: \emph{ceci n'est pas une panthère rose}.

Fig.~\ref{main_result} (right) shows that our stimuli maybe more similar to bunnies than to pink panthers. These MSE results confirm our paraphrase of Magritte's observation:
\emph{the digital photograph of our original drawing definitely is not a pink panther}.

\section{4. Acknowledgements}

This work was partially funded by the EU/FEDER/MICINN grants DPI2017-89867-C2-2-R and PID2020-118071GB-I00.

\bibliographystyle{jovcite}


\end{document}